\documentclass[10pt,twocolumn,letterpaper]{article}

\usepackage{wacv}
\usepackage{times}
\usepackage{graphicx}
\usepackage{bm}
\usepackage{amsmath,amssymb} 
\DeclareMathOperator*{\argmax}{arg\,max}
\usepackage{color}
\usepackage{multirow}
\usepackage{tabularx}
\usepackage[pagebackref=true,breaklinks=true,letterpaper=true,colorlinks,bookmarks=false]{hyperref}
\usepackage{cite}
\usepackage{float}

\wacvfinalcopy

\def\wacvPaperID{207} 

\ifwacvfinal\fi
\begin{document}

\title{Online Multi-Object Tracking with Instance-Aware Tracker and Dynamic Model Refreshment} 

\author{}
\author{Peng Chu, Heng Fan, Chiu C Tan, and Haibin Ling \\
Temple University, Philadelphia, PA USA\\
{\tt\small \{pchu, hengfan, chiu.tan, hbling\}@temple.edu}
}

\maketitle
\ifwacvfinal\thispagestyle{empty}\fi
\begin{abstract}

Recent progresses in model-free \emph{single object tracking} (SOT) algorithms have largely inspired applying SOT to \emph{multi-object tracking} (MOT) to improve the robustness as well as relieving dependency on external detector. However, SOT algorithms are generally designed for distinguishing a target from its environment, and hence meet problems when a target is spatially mixed with similar objects as observed frequently in MOT. To address this issue, in this paper we propose an instance-aware tracker to integrate SOT techniques for MOT by encoding awareness both within and between target models. In particular, we construct each target model by fusing information for distinguishing target both from background and other instances (tracking targets). To conserve uniqueness of all target models, our instance-aware tracker considers response maps from all target models and assigns spatial locations exclusively to optimize the overall accuracy. Another contribution we make is a dynamic model refreshing strategy learned by a convolutional neural network. This strategy helps to eliminate initialization noise as well as to adapt to the variation of target size and appearance. To show the effectiveness of the proposed approach, it is evaluated on the popular MOT15 and MOT16 challenge benchmarks. On both benchmarks, our approach achieves the best overall performances in comparison with published results.

\end{abstract}

\begin{figure*}[!h]
	\centering
	\includegraphics[width=0.9\linewidth]{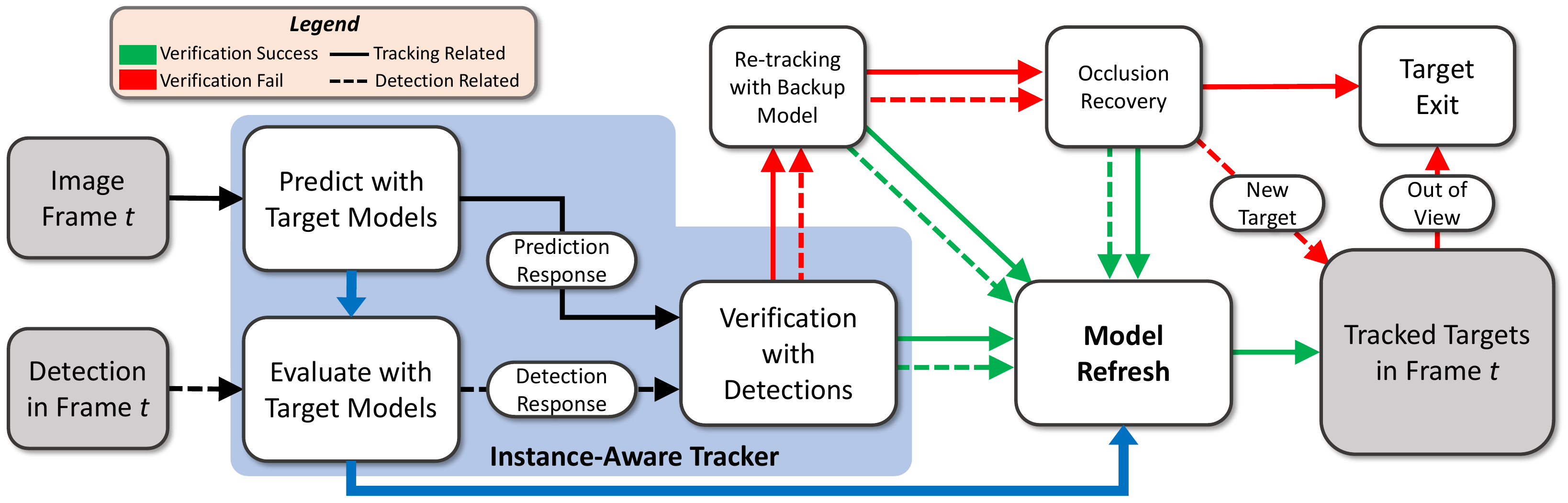}
	\caption{Overview of our instance-aware tracker based tracking system. }
	\label{fig:system_overview}
\end{figure*}

\section{Introduction}

Tracking multiple objects in video is critical for many applications, ranging from vision-based surveillance to autonomous driving. A popular solution to Multiple Object Tracking (MOT) is the tracking-by-detection strategy, in which, detections from an external detector on each frame are associated and connected to form target trajectories in either online or offline batch mode. With recent progress on object detector, tracking-by-detection has been successful in multiple domains~\cite{leal2014learning,bae2014robust,lenz2015followme,milan2016multi,wu2012coupling,brendel2011multiobject,berclaz2011multiple,leibe2007coupled,jiang2007linear}. However, separation of detection from tracking keeps detector inaccessible to the frame-to-frame correlation information which identifies the difference between object detection in still images and in videos. Moreover, the dependence on detection becomes a major limitation in complex scenes due to the degraded detection reliability caused by large size variation and partial occlusion of targets.


MOT, on the other hand, can be viewed as a generalized Single Object Tracking (SOT) problem where target locations are estimated from multiple SOT tracking models. Significant improvement has been achieved in recent SOT approaches which are efficient and robust in complex scenes~\cite{henriques2015high,li2014scale,bertinetto2016staple,fan2017sanet,fan2017parallel}. However, even with a proper target management mechanism, directly applying multiple SOT trackers simultaneously to track multiple targets still experiences various difficulties.

A SOT tracker usually allows certain generalisability to capture appearance changes of target. In the MOT context, however, multiple similar targets may appear in the searching area of a SOT tracker. Such targets from the same category (e.g., pedestrian) often share similar appearance or shape that may confuse traditional SOT trackers. When this happens, SOT trackers for multiple targets may easily drift and even end up tracking the same target. Moreover, since SOT trackers depend heavily on the model learned at the first frame, steady tracking of a model-free SOT tracker requires groundtruth bounding box of target at the first frame to correctly distinguish the target from its background. In the current MOT framework, all target candidates are provided by a real detector which usually yields considerable noise in both target location and scale.

In this work, we propose using \emph{instance-aware} (IA) tracker to both harvest the merit of SOT techniques and address the above issues in MOT. In addition to distinguishing a target from background as ordinary SOT tracker, our IA tracker tracks with the awareness of all other instances and their tracking models, which often means different targets of the same category. We implement such awareness in both target and global level. In scope of each target, we formulate the IA tracker in the efficient kernel correlation filter framework, while fusing features that tell a target from both background and other instances. This way, an IA target model is entitled with the awareness of differences between instances thus enhances response to its own target while suppressing responses to other similar targets. In global scope, generated response maps for all target models are used integrally to predict the target locations for a new coming frame. Awareness between targets models is treated as an optimization problem to maximize the overall response that each target is tracked exclusively by only one target model. A detection verification mechanism is proposed to solve the global optimization problem efficiently by incorporating detections from detectors and predictions from target models. And instead of updating model gradually, identity of a target model in proposed method is reinforced through a model refreshing mechanism, which is adaptively learned via a convolutional neural network.

Our contributions are mainly two-fold:
\begin{itemize}
\item We propose a novel instance-aware tracker to effectively integrate SOT in MOT. By being instance-aware inherently and mutually, target models significantly improve their capability to solve the ambiguity of similar targets in neighborhood.

\item We propose an adaptive model refreshment strategy to further improve the reliability of SOT in MOT context. 
\end{itemize}
To show the effectiveness of the proposed approach, it is evaluated on the popular MOT15 and MOT16 challenge benchmarks. On both benchmarks, our approach achieves the best overall performances in comparison with published results.


\section{Related Work}
\label{sec_relate_work}
Recent works on MOT primarily focuses on the tracking-by-detection principle. Most of these methods can be roughly categorized into two groups. The first group treat MOT as an offline global optimization problem that uses frame observation from both previous and future to estimate the current status of targets~\cite{ban2016tracking,pirsiavash2011globally,kim2016cdt,manen2016leveraging,mclaughlin2015enhancing}. These methods usually focus on data association based methods such as Hungarian algorithm~\cite{bewley2016simple,huang2008robust}, network flow~\cite{zamir2012gmcp,zhang2008global} and multiple hypotheses tracking~\cite{kim2015multiple}. Their performance heavily depends on the quality of detections from external detector. Different from these methods, our approach learns tracking model for each target to search and predict locations of next frame online. Detections in our approach are only used for model uniqueness verification and model refreshing.

The second group only needs observations till to the current frame to online estimate target status~\cite{yang2012online,fagot2015online,xiang2015learning,sadeghian2017tracking,chu2017online,yan2012track,yoon2015bayesian}. In \cite{xiang2015learning}, MOT is formulated as a Markov decision process with a policy estimated on the labeled training data. \cite{sadeghian2017tracking} extends the work \cite{xiang2015learning} to use deep CNN and LSTM to encode long-term temporal dependencies by fusing clues from motion, interaction and person re-identification model. Chu, et al.~\cite{chu2017online} use a dynamic CNN-based framework with a learned spatial-temporal attention map to handle occlusion, where CNN trained on ImageNet is used for pedestrian feature extraction. Yan et al.~\cite{yan2012track} gather target candidates from both detector and independent SOT trackers and select the optimal candidates through an ensemble model. Our approach differs from these methods by adding awareness between SOT trackers and dynamically refreshing model to eliminate possible noise in model initialization.

\section{System Overview}
\label{sec_overview}
For the $t$-th frame, our tracking system takes image frame and detections from an external detector as input, as shown in Fig.~\ref{fig:system_overview}. Target models of instance-aware tracker are used to predict target locations independently and estimate scores for each detection. A detection verification process is applied to assign each spatial candidate exclusively to only one tracked target and verify the uniqueness of their target model as detailed in Sec.~\ref{sec:tekcf} and  Sec.~\ref{det_verif}. Model of verified target will be refreshed if assigned detection enclosing target better than its model prediction. Unverified targets and detections will be matched again using backup models to recover from incorrect refreshment. These components are explained in Sec.~\ref{model_refresh}. Further, unpaired predictions and detections are passed into occlusion handling. Final unverified targets will exit when they have not been verified for some continuous frames. Unpaired detections will be added as new targets as described in Sec.~\ref{targ_m}.

\section{Methodology}
\label{sec_method}
\subsection{Problem Formulation}

Following the tracking-by-detection paradigm, online MOT can be formulated as an optimization problem, at frame $t$, the set of $N^t$ target locations $\hat{\bm{X}}^{t} = \big\{ \hat{x}^{t}_{i}\big\}^{N^t}_{i=1} $ in current image $I^t$ are chosen from $M^t$ candidates in set $\bm{O}^t = \big\{x^{t}_j\big\}^{M^t}_{j=1}$ to maximize a score:

\begin{equation}
\hat{\bm{X}}^{t} = \argmax_{\bm{X}^t \subset \bm{O}^t} f(I^t, \bm{X}^t; \mathbf{a}^{t}, \bm{W}^{t-1}).
\end{equation}
\begin{equation}
\text{s.t.} \quad \sum_i a^{t}_{ij} \le 1, a^{t}_{ij} \in \{0, 1\}.
\label{eqn:st}
\end{equation}
The parameter $\mathbf{a}^{t} = \big\{a^{t}_{ij}\in\{0,1\}\big\}$ indicates the association between the $i$-th tracked targets in $\hat{\bm{X}}^{t - 1}$ at frame $t - 1$ and the $j$-th candidate location in $\bm{X}^{t} $ at frame $t$. $a^{t}_{ij} = 1$ if $\hat{x}^{t - 1}_{i}$ is associated with $x^{t}_{j}$, and $a^{t}_{ij} = 0$ otherwise. Each candidate can only be assigned to at most one tracked target. $\bm{W}^{t} = \big\{\mathbf{w}^{t}_i\big\}^{N^t}_{i=1}$ is the set of parameters to model each target, which is usually learned through a training procedure using the appearance or location information of target.


The objective function $f(\cdot)$ measures the overall quality of the tracking results for all targets at frame $t$, defined as below
\begin{equation}
\label{f}
f(I^t, \bm{X}^t) = \sum_{ij} a^{t}_{ij} g_i(I^t, x^{t}_j; \mathbf{w}^{t - 1}_i).
\end{equation}
The set of functions $g_i(.)$ can be interpreted as the objective function for tracking single target such that $ g_i(I^t, x^{t}_j; \mathbf{w}^{t - 1}_i)$ assigns a score to the $j$-th candidate location $x^{t}_j$ on $I^t$ according to the $i$-th model parameter $ \mathbf{w}^{t - 1}_i \in \bm{W}^{t-1}$. The model parameters should be determined by previous images and target locations up to frame $t-1$ .

Solving the online MOT problem, therefore, is to solve $\mathbf{a}^{t}$ and $g_i(.)$ for each frame.

\subsection{Instance-Aware Tracker}
\label{sec:tekcf}

\begin{figure}[!t]
	\centering
	\includegraphics[width=0.996\linewidth]{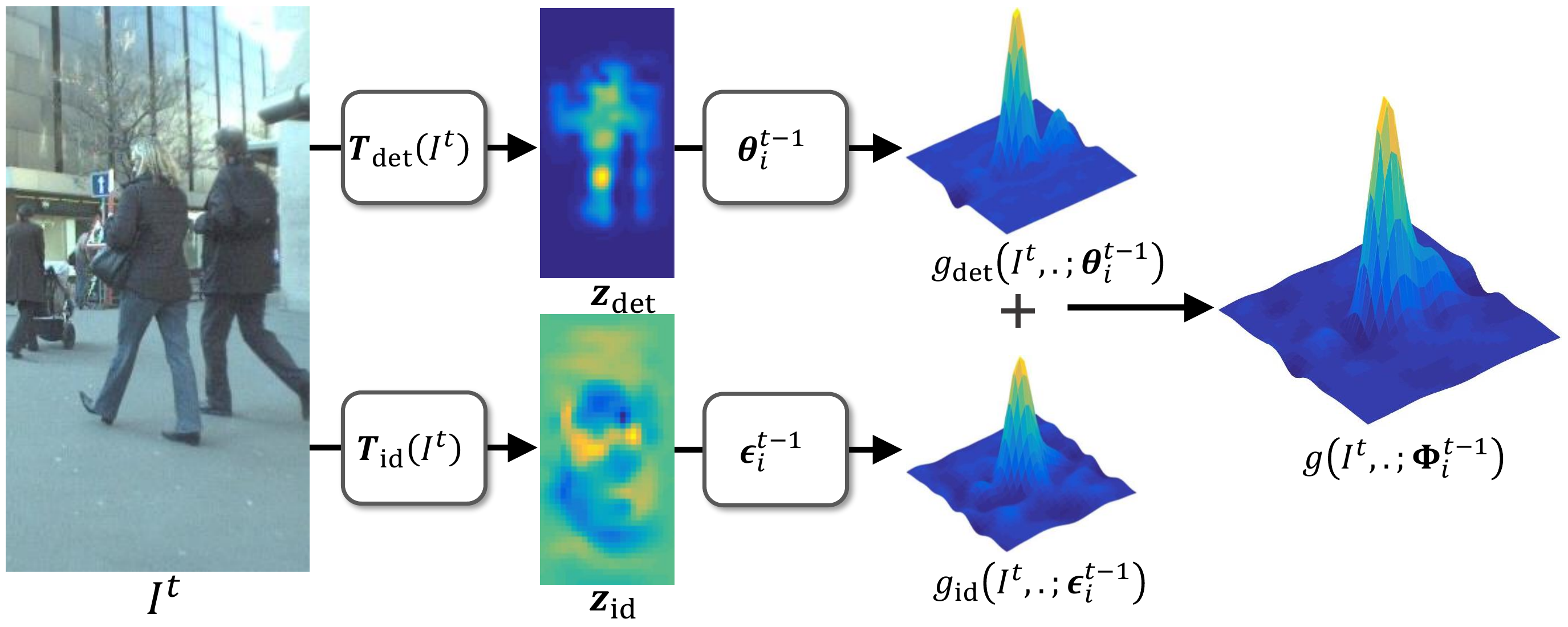}
	\caption{Illustration of target level instance-awareness: discriminate targets from background and discriminate different targets. $\mathbf{z}_{\mathrm{det}}$ and $\mathbf{z}_{\mathrm{id}}$ are feature maps visualized by accumulating values in all channels. }
	\label{fig:feat}
\end{figure}

\begin{figure*}[!t]
	\centering
	\includegraphics[width=0.87\linewidth]{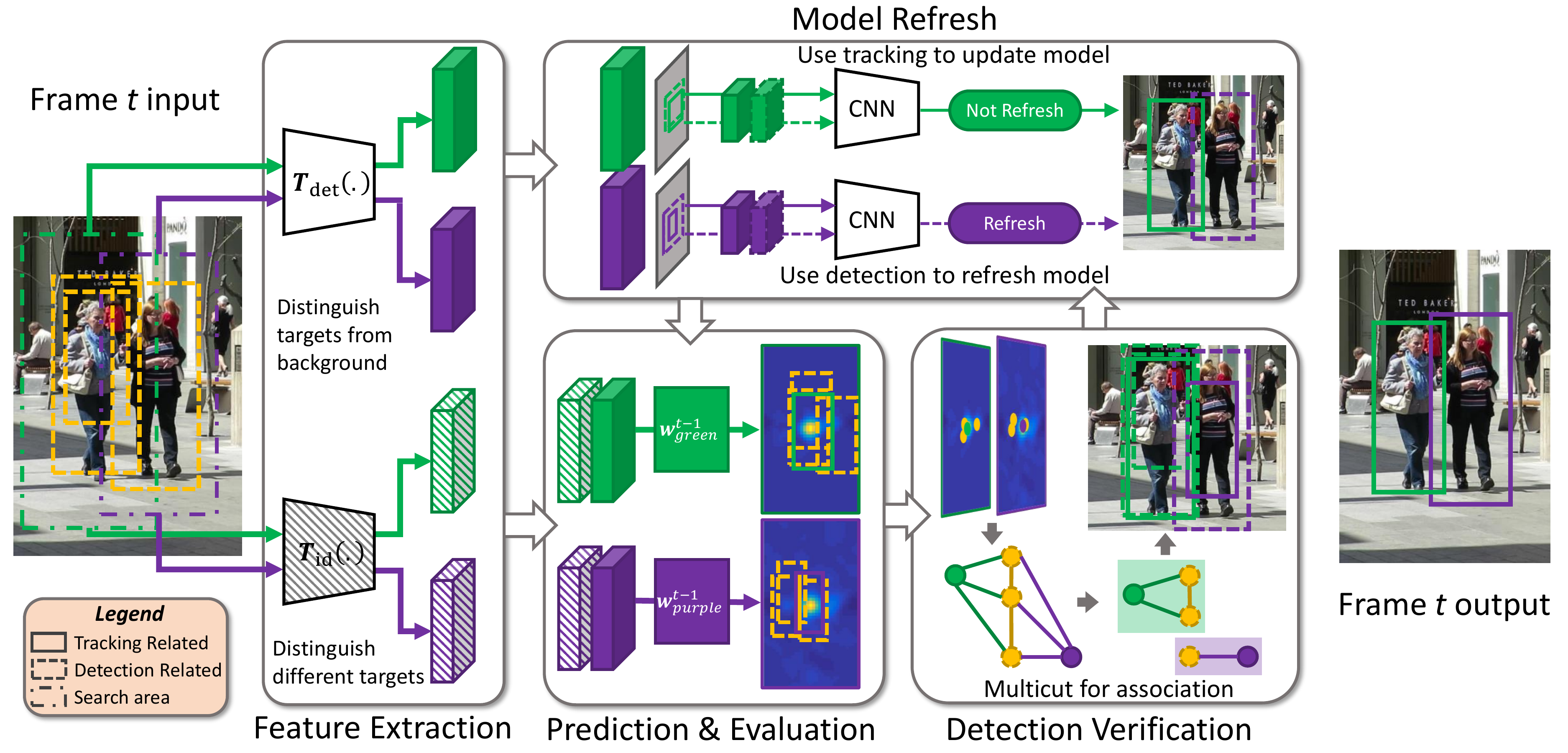}
	\caption{Tracking multiple objects by instance-aware tracker with model refreshment.}
	\label{fig:tracking}
\end{figure*}

We propose to use Instance-Aware (IA) tracker to solve $\mathbf{a}^{t}$ and $g_i(.)$ in two levels. For each single target, objective function $g_i(I^t, x^{t}_j)$ is learned to only assign a high score to its own target while returns low scores for both background and other instances. As for global, $\mathbf{a}^{t}$ is solved to associate each spatial location $x^{t}_j$ on $I^t$ exclusively to only one target referring those scores from all targets.

We start from the objective function $g_i(I^t, x^{t}_j)$. Ordinary SOT methods focus on distinguishing target from background and allow certain variations to handle appearance change of target, which makes tracker insensitive to distracters that are apparently similar to target. Thus, directly adopting SOT methods for MOT causes the trackers easily drifting to wrong targets. In this work, we treat the problem of tracking single target in MOT context as two sub-problems: i) to distinguish targets from background; and ii) to model the difference between targets. The objective function can be rewritten as
\begin{equation}
g_i(I^t, x^{t}_j) = g_{\mathrm{det}}(I^t, x^{t}_j; \bm{\theta}^{t-1}_i) + g_{\mathrm{id}}(I^t, x^{t}_j; \bm{\epsilon}^{t-1}_i),
\label{eqn:single_obj}
\end{equation}
where $\bm{\theta}^{t-1}_i$ and $\bm{\epsilon}^{t-1}_i$ are the two model parameters for the $i$-th target focusing on each of the problems mentioned above, therefore $ g_{\mathrm{det}}(I^t, x^{t}_j; \bm{\theta}^{t-1}_i)$ estimates the score for location $x^{t}_j$ containing one of the targets using model parameter $\bm{\theta}^{t-1}_i$, $g_{\mathrm{id}}(I^t, x^{t}_j; \bm{\epsilon}^{t-1}_i)$ evaluates the similarity for object at $x^{t}_j$ referring to the $i$-th target using model $\bm{\epsilon}^{t-1}_i$. Benefit of the separation is that for some target categories each of the sub-problems already has well-founded methods and datasets for model learning. For example, in the case of tracking multiple pedestrian, the first sub-problem is pedestrian detection and the second one is person re-identification, and both have large scale datasets such as MSCOCO, CUHK~\cite{li2014deepreid}.

We focus on the Ridge Regression form of $g_{\mathrm{det}}(.)$ and $g_{\mathrm{id}}(.)$, in which, the functions share the form of $g(\mathbf{z};\mathbf{\Phi}) = \mathbf{\Phi}\mathbf{z}^T$ with $\mathbf{z}$ for regression input and $\mathbf{\Phi}$ for learnable parameter. Then the objective function in Eq.~\ref{eqn:single_obj} can be rewritten as
\begin{equation}
\begin{split}
g_i(I^t, x^{t}_j) &= \bm{\theta}^{t-1}_i \bm{T}_{\mathrm{det}}(I^t, x^{t}_j)^T + \bm{\epsilon}^{t-1}_i \bm{T}_{\mathrm{id}}(I^t, x^{t}_j)^T \\
&= \mathbf{\Phi}^{t - 1}_i \Big[\bm{T}_{\mathrm{det}}(I^t, x^{t}_j), \bm{T}_{\mathrm{id}}(I^t, x^{t}_j) \Big]^T
\end{split}
\end{equation}
where $\bm{T}_{\mathrm{det}}(., x^{t}_j)$ and $\bm{T}_{\mathrm{id}}(., x^{t}_j)$ are image transformations centering at $x^{t}_j$, $[.]$ is channel-wise concatenation, $\mathbf{\Phi}^{t - 1}_i$ is the combined model parameter for the $i$-th target. An illustration is shown in Fig.~\ref{fig:feat}. 

Solving $\mathbf{\Phi}^{t - 1}_i$ online usually involves carefully designed strategies for positive and negative sample collection. Kernel Correlation Filter (KCF) tracker proposed in \cite{henriques2015high} solve this problem efficiently in Fourier domain by combining circulant matrices and kernel trick. Following this formulation, model parameter of the $i$-th target at frame $t$ is obtained by $ \mathbf{\tilde{w'}}^{t}_i =  \frac{\mathbf{\tilde{y}}}{ \mathbf{\tilde{k}}^{\mathbf{z}^t\mathbf{z}^t} + \lambda}$,
where $\mathbf{z}_{t} = \Big[\bm{T}_{\mathrm{det}}(I^t),\bm{T}_{\mathrm{id}}(I^t) \Big]$ is the feature map, $\mathbf{k}^{\mathbf{z}^t\mathbf{z}^t}$ is defined as kernel correlation in \cite{henriques2015high}, $ \mathbf{\tilde{y}}=\mathcal{F}(\mathbf{y})$ is Discrete Fourier Transform (DFT) of regression labels. If considering a SOT context where only one target presents, an predicted location  $^{P}\!\hat{{x}}^{t}_{i}$ using the $i$-th target model then can be estimated as
\begin{equation}
\label{prediction}
^{P}\!\hat{{x}}^{t}_{i} = \argmax_{x^{t}_j \in \bm{O}^t} \mathcal{F}^{-1}(\mathbf{\tilde{k}}^{\mathbf{z}^{t-1}\mathbf{z}^t} \odot \mathbf{\tilde{w}}^{t - 1}_i)  \bigr\vert_{=x^{t}_j}.
\end{equation}

Now we apply the objective function of tracking a single target to MOT context by combining Eq.~\ref{f} and Eq.~\ref{prediction}. Given the set of $\big\{x^{t}_j\big\}$, the objective function of IA tracker subject to constrain in Eq.~\ref{eqn:st} is defined as
\begin{equation}
f(I^t, \bm{X}^t)=\max_{a_{ij} } \sum_{ij} a^{t}_{ij} \mathcal{F}^{-1}(\mathbf{\tilde{k}}^{\mathbf{z}^{t-1}\mathbf{z}^t} \odot \mathbf{\tilde{w}}^{t - 1}_i)  \bigr\vert_{=x^{t}_j}
\label{eqn:graph_org}
\end{equation}

The core idea of IA tracker can be explained as follow. In SOT version of KCF tracker, prediction is the spatial location strongest responding on response map given by  $\mathcal{F}^{-1}(\mathbf{\tilde{k}}^{\mathbf{z}^{t-1}\mathbf{z}^t} \odot \mathbf{\tilde{w}}^{t - 1}_i)$. While in MOT, each spatial location has multiple responses generated by different  target models of frame $t-1$ as shown in Fig.~\ref{fig:tracking}. And to further confirm each target is tracked exclusively by only one target model, we make use of the spatial exclusive assumption that no two or more targets can occupy the same position on image at the same time (on image frame only, not considering real 3D space). Thus, a global optimization in Eq.~\ref{eqn:graph_org} is employed to maximize the overall response subject to the spatial exclusive constrain defined in Eq.~\ref{eqn:st} that each spatial location on image belongs to at most one target. 
Notice that, in calculation, $\mathbf{z}^{t}$ usually covers the search area of a target model only, where it should be written as $\mathbf{z}^{t}_i$ and the actual coordinate of $x^{t}_j$ in $\mathbf{z}^{t}_i$ should also be converted accordingly. 



\subsection{Detection Verification}
\label{det_verif}

Solving the optimization problem in Eq.~\ref{eqn:graph_org} for all spatial locations on image frame, e.g. each pixel, is computationally impractical. Ideally, a subset whose elements are complete and spatial exclusive is preferred. Prediction $^{P}\!{x}^{t}_{j} \in \bm{P}^t$ from Eq.~\ref{prediction} contains all possible locations for all targets, but these locations may have potential spatial conflicts. Detections $^{D}\!{x}^{t}_{j} \in \bm{D}^t$ from a category detector are spatial exclusive but not complete due to the possible false negative. We use the combination of detection $^{D}\!{x}^{t}_{j} \in \textbf{D}^t$ and predictions $^{P}\!{x}^{t}_{j} \in \textbf{P}^t$ as the candidate locations set $\bm{O}^t = \bm{D}^t \cup \bm{P}^t.$ Result candidate set is complete but only partially spatial exclusive.  Therefore, we propose a detection verification mechanism to solve $\mathbf{a}^{t}$ for all targets leveraging the limited spatial exclusive information provided by $\bm{D}^t$.


If a graph $G(\bm{V}, \bm{E})$ is created on $\bm{V} = \hat{\bm{X}}^{t - 1} \cup \bm{O}^t$ and $\bm{E}$ which are the edges between vertexes in $\bm{V}$, the optimization problem in Eq.~\ref{eqn:graph_org} with constrain in Eq.~\ref{eqn:st} can be reformed as a graph multicut problem minimizing cost:

\begin{equation}
\min_{c_{e} \in \{0, 1\}} \sum_{e \in E} c_{e} d_{e}
\label{eqn:graph}
\end{equation}
\begin{equation}
\text{s.t.} \quad \forall P_{uv} \in \mathcal{P} \quad \forall e \in P_{uv} : c_e \le \sum_{e' \in P_{e} \slash \{ e \}} c_{e'},
\label{eqn:graph_st}
\end{equation}
where $u,v \in \bm{V}$, $c_e$ is the binary label indicating if $e \in \bm{E}$ is a cutting edge, $d_e$ is the cost/reward associated to edge $e$, $P_{uv}$ is the set of path from $u$ to $v$, $e$ is the edge between $u$ and $v$. Solving Eq.~\ref{eqn:graph} and Eq.~\ref{eqn:graph_st} in the context of MOT is to find the subgraphs, in which, candidate locations belonging to the same target are connected while belonging to different targets are separated by cutting edges as shown in Fig. \ref{fig:tracking}.

After optimization, each $^{D}\!{x}^{t}_{j}$ is assigned to one of the tracked target $ \hat{x}^{t - 1}_{i} \in \hat{\bm{X}}^{t - 1} $. Verification of each target tracked exclusively by only one tracking models can be done by checking whether a tracked target assigned with detections. Due to the possible false negative and false positive generated by a real detector, verification can only be conducted every $T_V$ frames to confirm the uniqueness of target model in long-term. Particular, if a tracked target has not been assigned with any detection for continuous $T_V$ frames, then its target model is likely tracking either a false positive target or a target shared with other models. Increasing $T_V$, therefore, decreases awareness between target models since it allows each target model to track independently for more frames. $T_V$ also controls the dependency on external detection and can be adjusted to adapt different detection qualities. Detailed parameter choice and discuss for $T_V$ are described in Sec.~\ref{sec_tv}.

We employ a primal heuristic based approach proposed in \cite{keuper2015efficient} for solving Eq.~\ref{eqn:graph} with Eq.~\ref{eqn:graph_st}, where a set of transformation sequences are used to update the bi-partitions of a subgraph. Specifically, cost of each edge in Eq.~\ref{eqn:graph} is calculated as:
\begin{equation}
d_{e} \doteq d_{uv} = \begin{cases}
g_{i}(I^t, x^{t}_j), &\mathrm{if~} u \in \hat{\bm{X}}^{t - 1}, v \in \bm{O}^t \\
\mathrm{IoU}(\bm{b}^{t}_j, \bm{b}^{t}_{j'}) , &\mathrm{if~}  u \in \bm{O}^t, v \in \bm{O}^t \\
-C , &\mathrm{if~}  u \in \hat{\bm{X}}^{t - 1}, v \in \hat{\bm{X}}^{t - 1} \\
\end{cases}
\end{equation}
where $\mathrm{IoU}(.)$ calculates the bounding box overlap ratio in term of Intersection over Union, $\bm{b}^{t}_{j}$ is the bounding box associated with $x^{t}_j$, and $C$ is large positive constant to ensure cutting between different tracked targets. The final equivalence between $a_{ij}$ and $c_e$ is defined as following
\begin{equation}
a_{ij} = \begin{cases}
\bar{c}_e = \bar{c}_{uv}, & \mathrm{if~} u \in \hat{\bm{X}}^{t - 1}, v \in \bm{O}^t \\
0, & \mathrm{otherwise}
\end{cases}
\end{equation}
where $\bar{c}_e$ stands for the logical negation.

\begin{table*}[!h]
	\small
	\begin{center}
		\caption{Tracking Performance on the MOT training set.}	
		\label{tab_tv}
		\begin{tabular}{cccccccccccc}
			\hline\hline
			\multicolumn{12}{c}{Venice-2}\\
			\hline
			$T_V$&0&1&2&3&4&5&6&7&8 &10 & 20 \\
			MOTA&27.0&30.9&32.7&33.7&34.4&32.4&33.1&33.4& 33.1 &  34.1 & 31.5\\
			FP & 647 &708 &773 &795 &824 &864 &889 &922 &942 &961   & 1237 \\
			FN& 4498 &4173 &3991 &3898 &3825 &3931 &3856 &3801 &3801 &3714    & 3622\\							
			\hline
			\multicolumn{12}{c}{MOT16-05}\\
			\hline
			$T_V$&0&1&2&3&4&5&6&7&8 & 10 & 20\\
			MOTA& 39.5&40.3&40.9&41.1&41.2 &42.3&42.3&42.2& 40.9 & 40.0& 37.8\\
			FP & 187  &231 &255 &281 &312 &324 &340 &355 &415 & 492 & 693\\
			FN & 3522 &3441 &3387 &3349 &3314 &3240 &3228 &3216 &3241 & 3222&3152\\			
			\hline\hline
		\end{tabular}
		
	\end{center}	
\end{table*}

\begin{table*}[!h]
	\small
	\begin{center}
		\caption{Tracking Performance on the MOT training set}
		\label{tab:ab}
		\begin{tabular}{lcccccccc }
			\hline\hline
			Method & MOTA$\uparrow$ & MOTP$\uparrow$ & MT$\uparrow$ & ML$\downarrow$ & FP$\downarrow$ & FN$\downarrow$ & IDS$\downarrow$ & Frag$\downarrow$\\
			\hline
			IA & 26.1 & 69.3 & 15.4\% & 26.9\% &  991  & 4250 &  39  & 42 \\
			IA+MR & 33.3 & 74.0  & 15.4\% & 34.6\% & 785 & 3939 &  36 &  53\\
			IA$-$DV+MR & -35.3 &  72.7 & 38.5\% &  19.3\% & 6734 & 2856 &  70  & 108 \\
			IA$-$TA+MR & 32.2 &  73.7 & 15.4\% &  38.5\% & 760 & 4039 &  43  & 53 \\
			Full & 34.4 & 74.1& 15.4\% &  30.8\%  &824 &3825 &36& 66\\			
			\hline\hline
		\end{tabular}
		
	\end{center}	
\end{table*}

\subsection{Model Refresh}
\label{model_refresh}
We train a CNN based classifier to determine whether to refresh the tracking model of a target using its assigned detection. Target model in ordinary SOT methods is initialized by target groundtruth in the first frame and is slowly and constantly updated.  While in MOT, models are initially learned from detections which contain considerable noise in location and scale. And when targets moving close to the camera, their scale also will change rapidly. Due to those reasons, models in MOT have to be refreshed frequently.

Specifically, feature maps centering at tracked target $^{P}\!\hat{x}^{t}_{i}$ and its assigned detection $^D\!x^{t}_{j}$ are extracted and stacked channel-wise to feed into a CNN based classifier. The CNN is to make comparison between the bounding boxes associated with $^{P}\!\hat{x}^{t}_{i}$ and $^D\!x^{t}_{j}$ on target enclosing. If the bounding box of $^{D}\!x^{t}_{i}$ encloses target better, $\mathbf{w}^{t}_i$ will be refreshed by re-calculating $\mathbf{w}^{t}_i$ using $^{D}\!x^{t}_{i}$. In tracking phase, we reuse features from $\bm{T}_{\mathrm{det}}(I^t)$ and adopt ROI Pooing to exact feature maps at specific locations, as show in Fig.~\ref{fig:tracking}

We adopt reinforcement learning to train the CNN classifier for the model refreshing policy. We update the classifier or the policy only when it makes a mistake. Suppose the tracker is tracking the $i$-th target in $t$-th frame. There are two types of mistake that can happen. i) Bounding box of $^{D}\!x^{t}_{i}$ encloses target better than $^{P}\!\hat{x}^{t}_{i}$ referring to groundtruth bounding box, but classifier chooses not to refresh $\mathbf{w}^{t}_i$. Then features at $^{D}\!x^{t}_{i}$ and $^{P}\!\hat{x}^{t}_{i}$ are concatenated and added to training set as positive samples. ii) Bounding box of $^{P}\!\hat{x}^{t}_{i}$ encloses target better than $^{D}\!x^{t}_{i}$, but classifier chooses to refresh $\mathbf{w}^{t}_i$. Concatenated features in those cases are added as negative samples. Each time the classifier makes a mistake, the CNN is trained through a constant number of iterations using online batches of size $B_N$, which contains the newly added sample and $B_N - 1$ samples randomly sampled from the rest training set. We keep updating the policy until all the targets in training set are successfully tracked.

In case of classifier making mistake at real tracking phase, we adopt a model backup mechanism. In frame $t$, if classifier chooses to refresh $\mathbf{w}^{t - 1}_i$ with a new $\mathbf{w}^{t}_i$, $\mathbf{w}^{t - 1}_i$ will be saved. In frame $t + 1$, if tracker with $\mathbf{w}^{t}_i$ cannot be assigned with a $^D\!x^{t + 1}_{j}$, the old model $\mathbf{w}^{t - 1}_i$ will be restored for tracking and verification one more time.

\subsection{Target Management}
\label{targ_m}
In this work, except for `Tracked' event, we also handle the `Occlusion', `Enter' and `Exit' events of targets.

\noindent\textbf{Occlusion} To recovery a target from occlusion, we train an SVM classifier to estimate if two locations $\hat{x}^{t'}_{i}$ and $^D\!x^{t}_{j}$ are containing the same target. We make a simple assumption that a detection $^D\!x^{t}_{j}$ not assigned to any tracked target in detection verification and re-tracking phase is either a new target or an existing target just finished occlusion. Occlusion recovery thus is to connect that detection with tracked target not assigned with detection. Suppose $i$-th target starts to be occluded in frame $t'$ and finishes occlusion in frame $t$, where $t - t' > 1$, $\mathbf{b_1} = (\xi_1, \zeta_1, \omega_1, \eta_1)$ is the bounding box associated with the first location specified by its $x$-coordinate, $y$-coordinate, width and height respectively, and $\mathbf{b}_2$ is for the second location.  We can calculate the following feature for estimation,
$$\Big[t-t', \frac{\xi_2 - \xi_1}{\bar{\eta}}, \frac{\zeta_2 - \zeta_1}{\bar{\eta}}, \frac{\eta_2 - \eta_1}{\bar{\eta}}, \mathrm{IoU}(\mathbf{b}_2, \mathbf{b}_1), \phi_{hist} \Big],$$
where $\bar{\eta}  = \frac{\eta_ 1 + \eta_ 2}{2}$ and $\phi_{hist}$ is histogram intersection of the two image patches bounded by $\mathbf{b}_1$ and $\mathbf{b}_2$. In the tracking phase, for those $^D\!x^{t}_{j}$ not assigned to any  $\hat{x}^{t - 1}_{i}$ and those $\hat{x}^{t'}_{i}$ not being assigned with any  $^D\!x^{t}_{j}$, the SVM classifier is used to estimate the matching possibility of each pair. Hungarian algorithm is employed to find the final matching pair.

\noindent\textbf{Target Enter} As mentioned above, if $^D\!x^{t}_{j}$ hasn't been assigned in any of the previous stages, $^D\!x^{t}_{j}$ is added to $\hat{\bm{X}}_{t}$ as a new target. 

\noindent\textbf{Target Exit} We adopt two criteria for target exit checking: i) Bounding box of $\hat{x}^{t}_{i}$ is out of view. ii) Target hasn't been assigned a detection for continuous $T_V$ frames.

\section{Experiments}
\label{exp}
We conduct three experiments on the popular MOT15 \cite{leal2015motchallenge} and MOT16 \cite{milan2016mot16} benchmarks to analyze our proposed approach and compare to prior works. The test set of MOT15 contains 11 sequences and MOT16 contains 7 sequences, where camera motion, camera angle, and imaging condition vary greatly. For each test sequence, a training sequence is provided which is captured in the similar settings. For both training and test set, detections from a real detector are provided. 

\begin{table*}[!t]
	\small
	\begin{center}
		\caption{Tracking Performance on the MOT15 benchmark test set. Best in bold.}
		\label{table:res15}
		\begin{tabular}{c|ccccccccc}
			\hline\hline
			Mode & Method & MOTA$\uparrow$ & MOTP$\uparrow$ & MT$\uparrow$ & ML$\downarrow$ & FP$\downarrow$ & FN$\downarrow$ & IDS$\downarrow$ & Frag$\downarrow$ \\
			\hline
			
			\multirow{6}{*}{\rotatebox{90}{\textbf{Offline}}}
			& TBD~\cite{geiger20143d} & 15.9 &	70.9&6.4\% &	47.9\% &14943&	34777&1939&	1963 \\
			&CEM \cite{milan2014continuous}& 19.3& 70.7& 8.5\%& 46.5\%& 14180 &34591 &813 &1023 \\
			&JPDA\_m \cite{rezatofighi2015joint}& 23.8 & 68.2& 5.0\%& 58.1\%& \textbf{4533}& 41873 &\textbf{404}& \textbf{792} \\
			& SiameseCNN \cite{leal2016learning}& 29.0& 71.2& 8.5\%& 48.4\%& 5160 &37798& 639& 1316 \\
			& MHT\_DAM \cite{kim2015multiple}& 32.4&71.8&16.0\% &43.8\% &	9064 &	32060&	435&826       \\
			& JMC \cite{keuper2016multi}&\textbf{35.6}&\textbf{71.9}&\textbf{23.2\%} &\textbf{39.3\%} &10580 &\textbf{28508} &457	&969	 \\
			\cline{1-10}
			
			\multirow{6}{*}{\rotatebox{90}{\textbf{Online}}}
			&RNN \cite{milan2017online}& 19.0	&71.0&	5.5\%&	45.6\% &11578&36706&	1490	&2081 \\
			&oICF \cite{kieritz2016online}& 27.1&70.0	&6.4\% &48.7\% &7594&36757&	\textbf{454}	&1660 \\
			& SCEA \cite{hong2016online}& 29.1	&71.1	&	8.9\% &	47.3\% &6060&	36912	&604&	\textbf{1182} \\
			& MDP \cite{xiang2015learning}& 30.3& 71.3& 13.0\%& 38.4\%& 9717 &32422 &680 &1500       \\
			& AP \cite{long2017online}&38.5& \textbf{72.6}& 8.7 \%& 37.4\%& \textbf{4005} &33203 &586 &1263 \\
			& proposed & \textbf{38.9}  & 70.6 & \textbf{16.6\%} & \textbf{31.5\%} & 7321	& \textbf{29501}	& 720 & 1440  \\
			
			\hline\hline
		\end{tabular}

	\end{center}
	
\end{table*}

\begin{table*}[!t]
	\small
	\begin{center}
		\caption{Tracking Performance on the MOT16 benchmark test set. Best in bold.}
		\label{table:res16}
		\begin{tabular}{c|ccccccccc}
			\hline\hline
			Mode & Method & MOTA$\uparrow$ & MOTP$\uparrow$ & MT$\uparrow$ & ML$\downarrow$ & FP$\downarrow$ & FN$\downarrow$ & IDS$\downarrow$ & Frag$\downarrow$ \\
			
			\hline
			\multirow{6}{*}{\rotatebox{90}{\textbf{Offline}}}
			& SMOT \cite{dicle2013way}& 29.7 &	75.2&	5.3\% &	47.7\% &	17426	&107552&	3108&	4483	 \\
			&CEM \cite{milan2014continuous}& 33.2&75.8&	7.8\% &	54.4\% &	6837&114322&	642	&731 \\
			&GMMCP \cite{dehghan2015gmmcp}& 38.1 &	75.8&	8.6\% &	50.9\% &	6607&	105315	&937&	1669 \\
			& MHT\_DAM \cite{kim2015multiple}& 45.8 &	76.3&	16.2\% &	43.2\% &	\textbf{6412}&	91758&	590&	781       \\
			& NOMT \cite{choi2015near}& 46.4	&76.6	&\textbf{18.3\%} &41.4\% &	9753&87565	&\textbf{359}	&\textbf{504}       \\
			& LMP \cite{tang2017multiple}&\textbf{48.8} &	\textbf{79.0} & 	18.2\% &	\textbf{40.1\%} &	6654&	\textbf{86245} &	481&595	 \\
			\cline{1-10}
			
			\multirow{5}{*}{\rotatebox{90}{\textbf{Online}}}
			&OVBT \cite{ban2016tracking}& 38.4	&75.4&	7.5\% &	47.3\% &	11517 &	99463 &	1321 &	2140	 \\
			& EAMTT	\cite{sanchez2016multi}& 38.8 &	75.1	&	7.9\% &	49.1\% &	8114&102452&965&	1657	 \\
			& oICF \cite{kieritz2016online}& 43.2 &	74.3&11.3\% &	48.5\% &	6651&	96515&	\textbf{381}&	1404       \\
			& AMIR \cite{sadeghian2017tracking}&47.2&	\textbf{75.8}&14.0\% &	41.6\% &	\textbf{2681}&92856&	774&1675 \\
			& proposed & \textbf{48.8} & 75.7   & \textbf{15.8\%} &	\textbf{38.1\%}&	5875&	\textbf{86567}&	906&	\textbf{1116}  \\
			\hline\hline 
		\end{tabular}
		
	\end{center}
	
\end{table*}

\subsection{Experiment Setting}

The proposed approach is implemented in MATLAB with Caffe and running on a desktop with 4 cores@3.60GHz CPU and a GTX1080 GPU. We use PAFNet proposed in \cite{cao2017realtime} for $\bm{T}_{\mathrm{det}}(.)$. PAFNet generates two feature maps at the end, where different human body parts and corresponding affinity field are highly responded. Two feature maps are concatenated along channel to form the output of $\bm{T}_{\mathrm{det}}(.)$. We use $\bm{T}_{\mathrm{det}}(.)$ to distinguish pedestrian from their background. PartNet proposed in \cite{zhao2017deeply} is adopt for $\bm{T}_{\mathrm{id}}(.)$. PartNet generate $L2$ normalized feature for person Re-Identification task, which is suitable for $\bm{T}_{\mathrm{id}}(.)$ to distinguish different pedestrians. Original PartNet outputs a feature vector for each input image. We remove its last global pooling layer and convert the last fully connected (FC) layer to convolutional layer to output feature map in reasonable dimensions. 

For each test sequence in MOT15 and MOT16 dataset, one or more similar sequences in training set are used to train a CNN classifier mentioned in Sec.~\ref{model_refresh} and a SVM classifier in Sec.~\ref{targ_m}. We adopt the partition method mentioned in \cite{xiang2015learning}. CNN classifier is consisted of one convolutional layer and one FC layer. When training the CNN classifier, $B_N = 32$ and 5 iterations with constant learning rate of 0.001 are used when CNN classifier makes mistake.

By implementing IA tracker and model refreshment with shared feature extraction as shown in Fig.~\ref{fig:tracking}, the average speed of proposed approach on MOT15 dataset is about 0.3 fps and 0.1 fps on MOT16 dataset. The average target densities on each frame of those two datasets are 10.6 and 30.8 for the test set. Proposed method achieves acceptable speed performance compared with other methods such as LMP (offline)\cite{tang2017multiple} at 0.6 fps and AMIR (online)\cite{sadeghian2017tracking} at 1.0 fps.

\noindent \textbf{Evaluation Metric} To evaluate the performance of proposed method, we employ the widely accepted CLEAR MOT metrics \cite{bernardin2008evaluating}, including multiple object tracking precision (MOTP) and multiple object tracking accuracy (MOTA) which is a cumulative measure that combines false positives (FP), false negatives (FN) and the identity switches (IDS). Additionally, we also report the percentage of mostly tracked targets (MT), the percentage of mostly lost targets (ML), and the number of times a trajectory is fragmented (Frag).

\subsection{Determine $T_V$}
\label{sec_tv}
Hyper-parameter $T_V$ in proposed approach is used to determine the maximum continuous frames that a target can be tracked without the verification from external detection. Setting of $T_V$ controls the strength of awareness between target models: As $T_V$ increasing, verification becomes less frequent, each target model tracks its target more independently, which is more equivalent with directly applying multiple SOT tracker for MOT; When decreasing $T_V$, proposed approach depends more on external detection and behaviors more like traditional tracking-by-detection approaches. As reflected in evaluation metrics, choice of $T_V$ controls the trade off between FP and FN. Higher $T_V$ allows tracker to continue more frames without the confirmation from detection, thus may introduce more FP. Lower $T_V$ requires frequent verification between tracker and detection, where tracking performance will heavily depend on detection quality, thus tracker may generate more FN when detection quality gets worse.

We test various of $T_V$ on the training dataset of MOT benchmark. The results of MOTA, FP and FN for Venice-2 from MOT15 and MOT16-05 from MOT16 are reported in Tab.~\ref{tab_tv}. In both sequences, starting at $T_V = 0$ where verification for every frame is required and increasing $T_V$, MOTA first increases then decreases due to the increasing FP in results. FP and FN gain their balance at $T_V = 4$ for Venice-2 and $T_V = 5$ for MOT16-05 where MOTA achieves best. We choose $T_V = 4$ for the rest of our experiments.

\begin{figure*}[!h]
	\small
	
	\centering
	
	\begin{tabular}
		{c@{\hspace{0.3mm}}c@{\hspace{.4mm}}c@{\hspace{.4mm}}c@{\hspace{.4mm}}}%
		\includegraphics[width=0.195\linewidth]{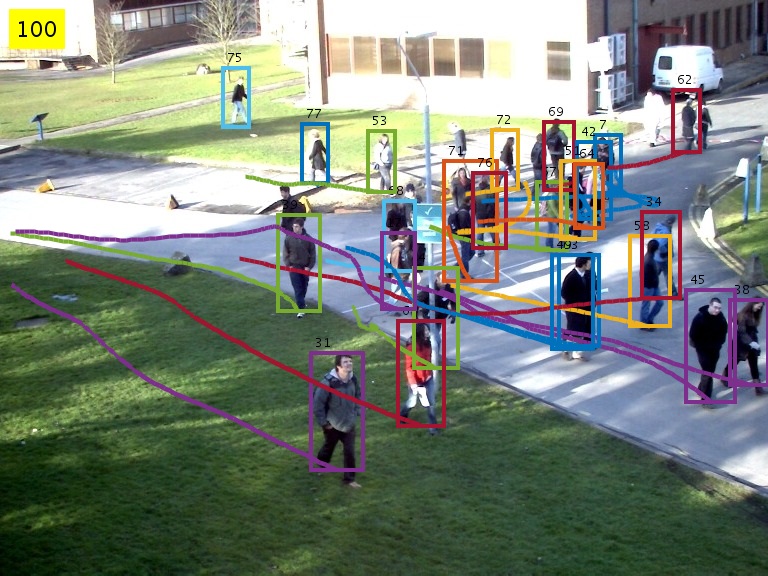}    & \includegraphics[width=0.26\linewidth]{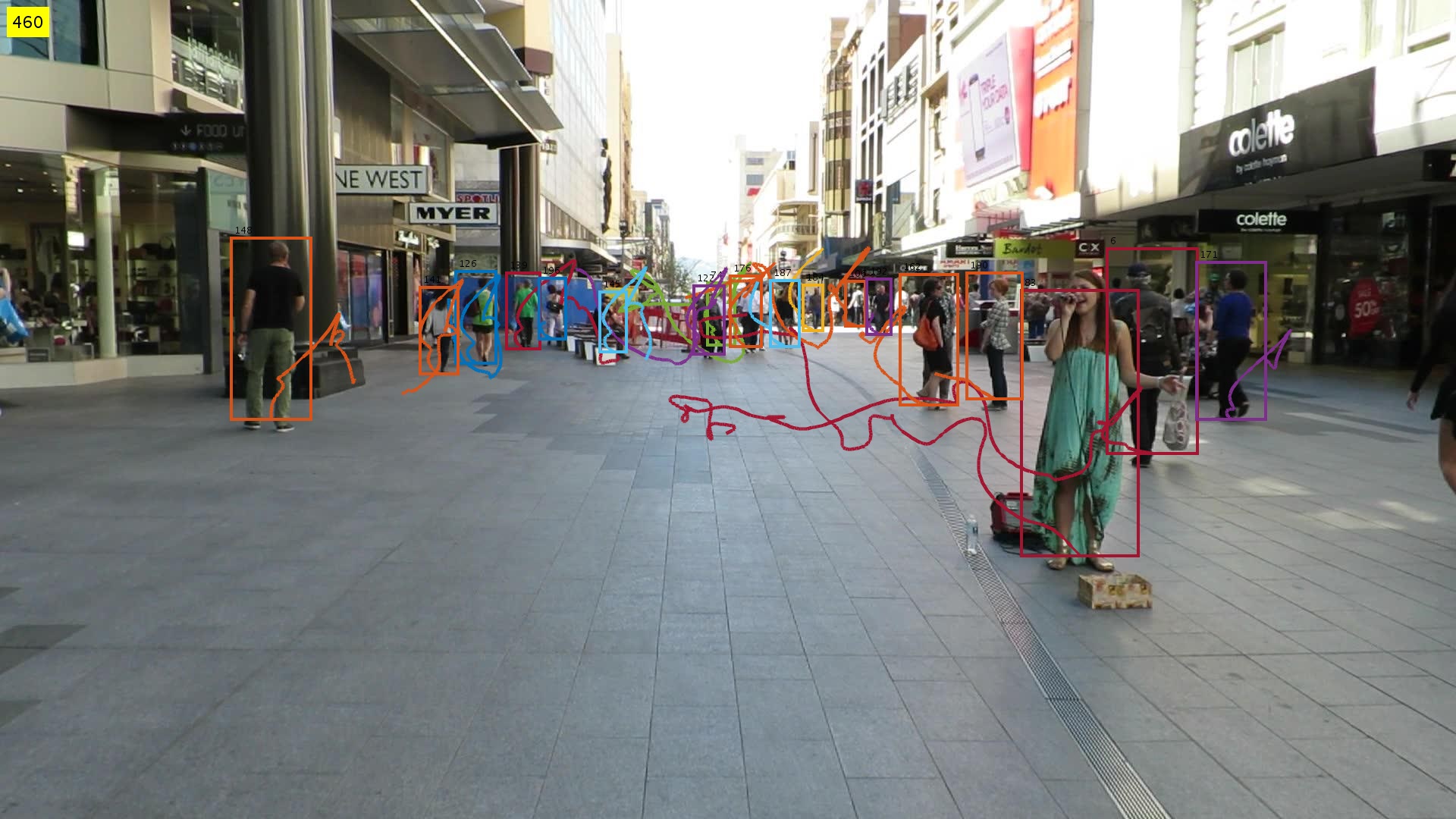}    & \includegraphics[width=0.26\linewidth]{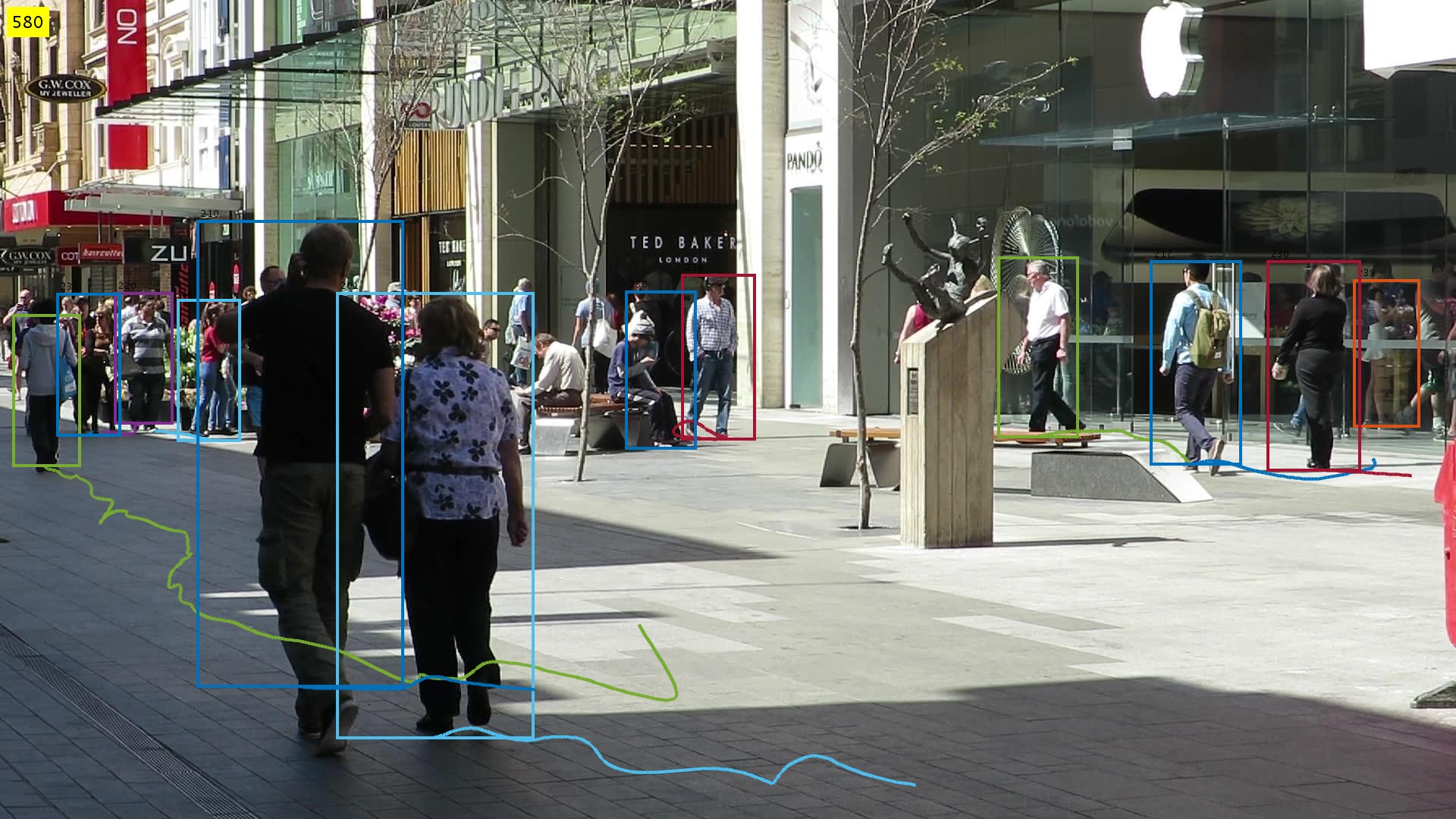}    &
		\includegraphics[width=0.26\linewidth]{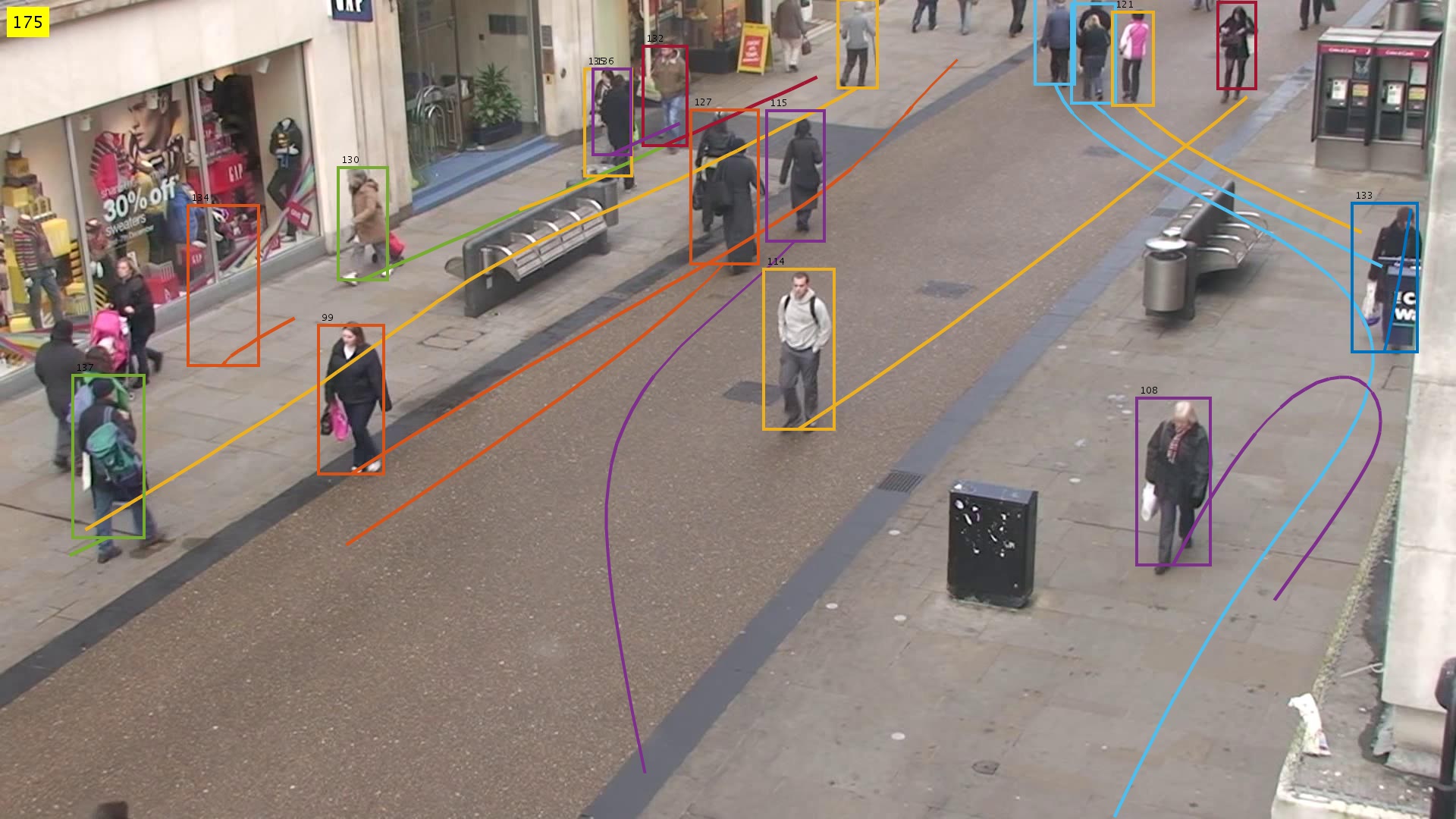}
		\\
		PETS09-S2L2& ADL-Rundle-1 & ADL-Rundle-3 & AVG-TownCentre
		\\
		\includegraphics[width=0.195\linewidth]{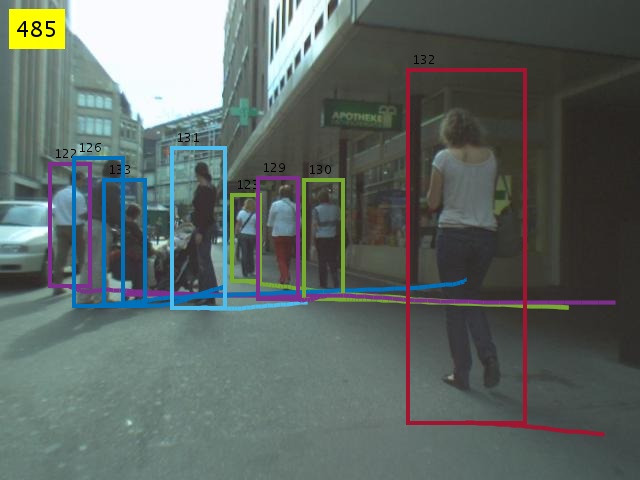}   & \includegraphics[width=0.26\linewidth]{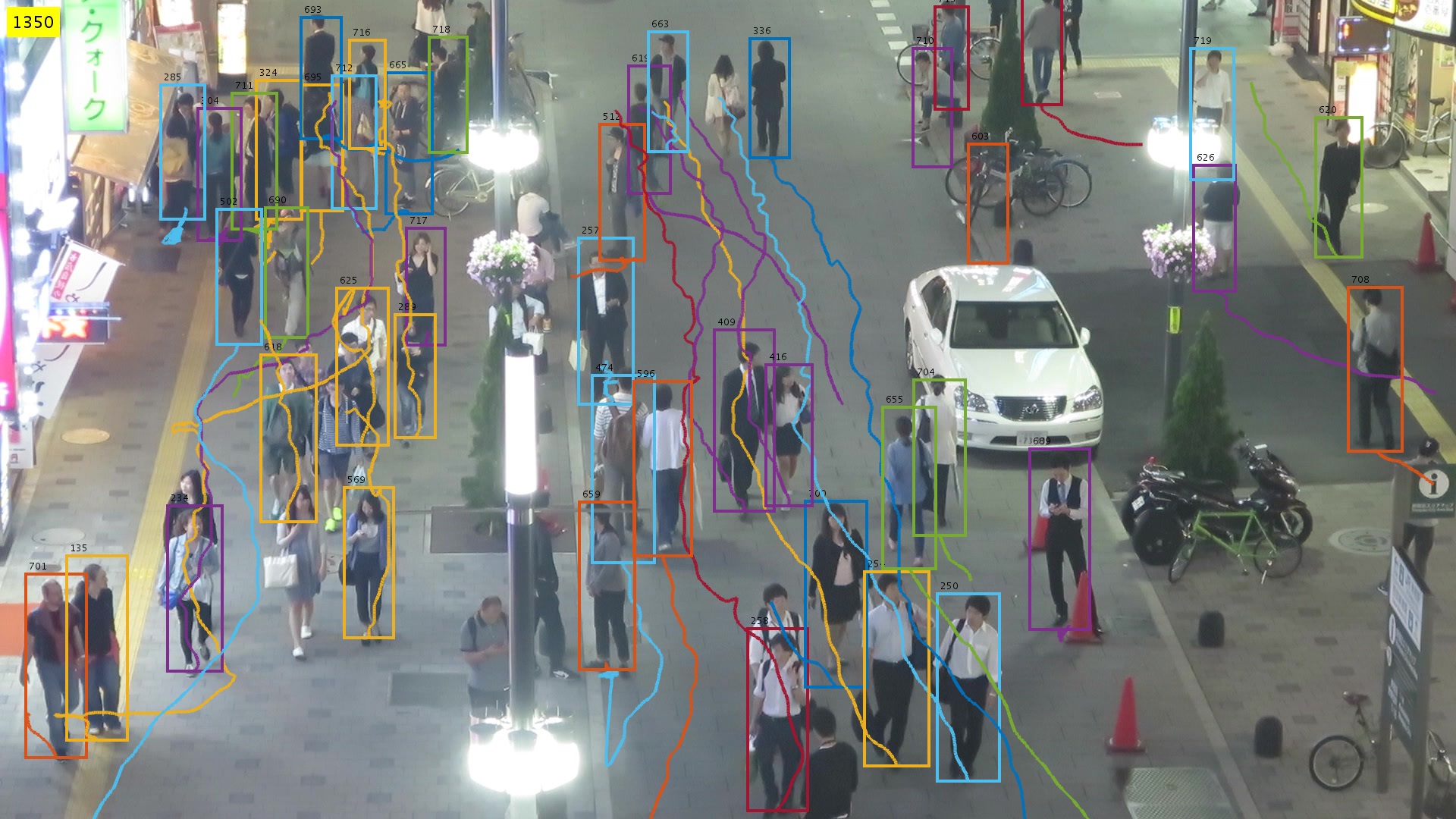}   & \includegraphics[width=0.26\linewidth]{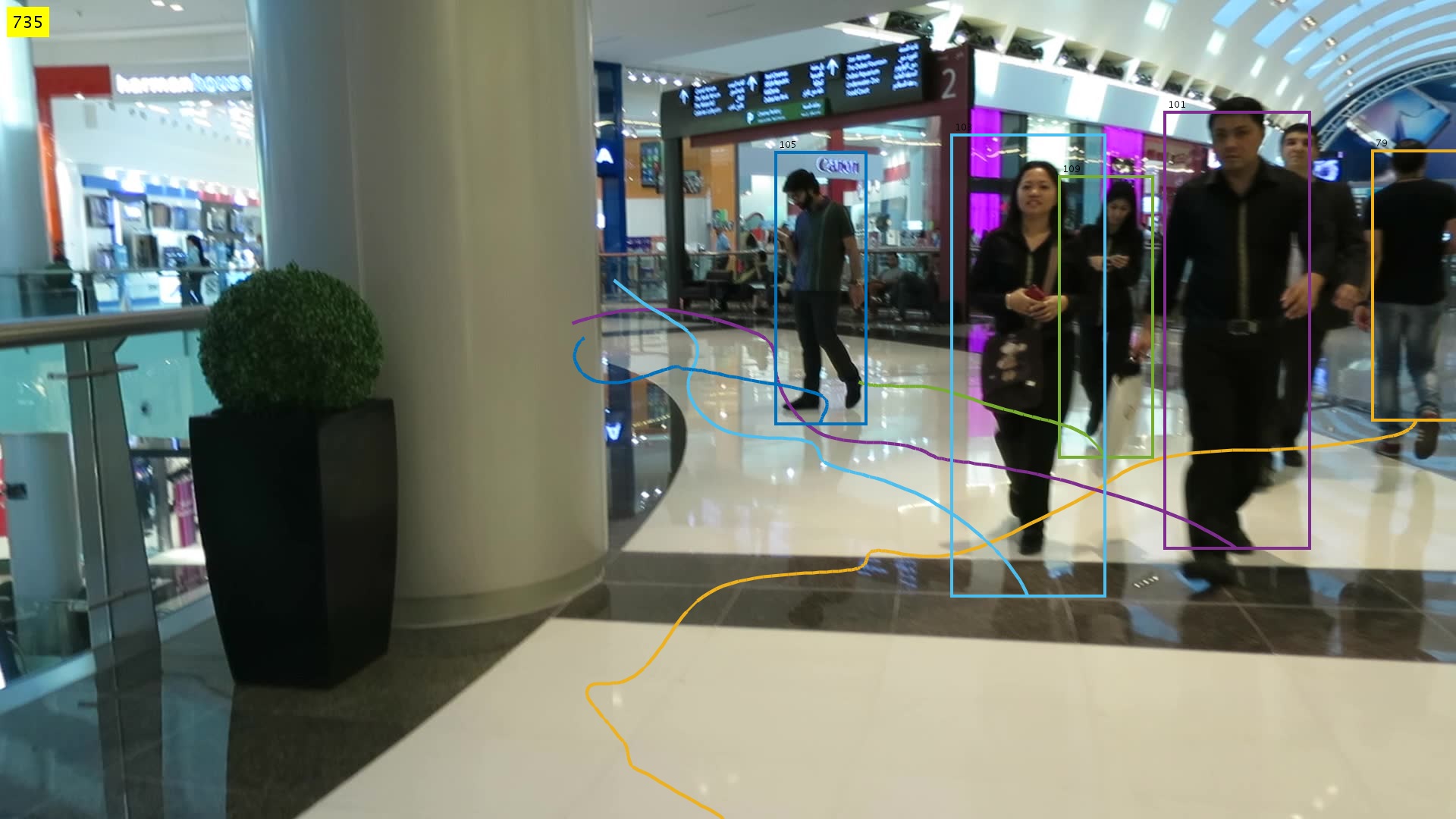}   &
		\includegraphics[width=0.26\linewidth]{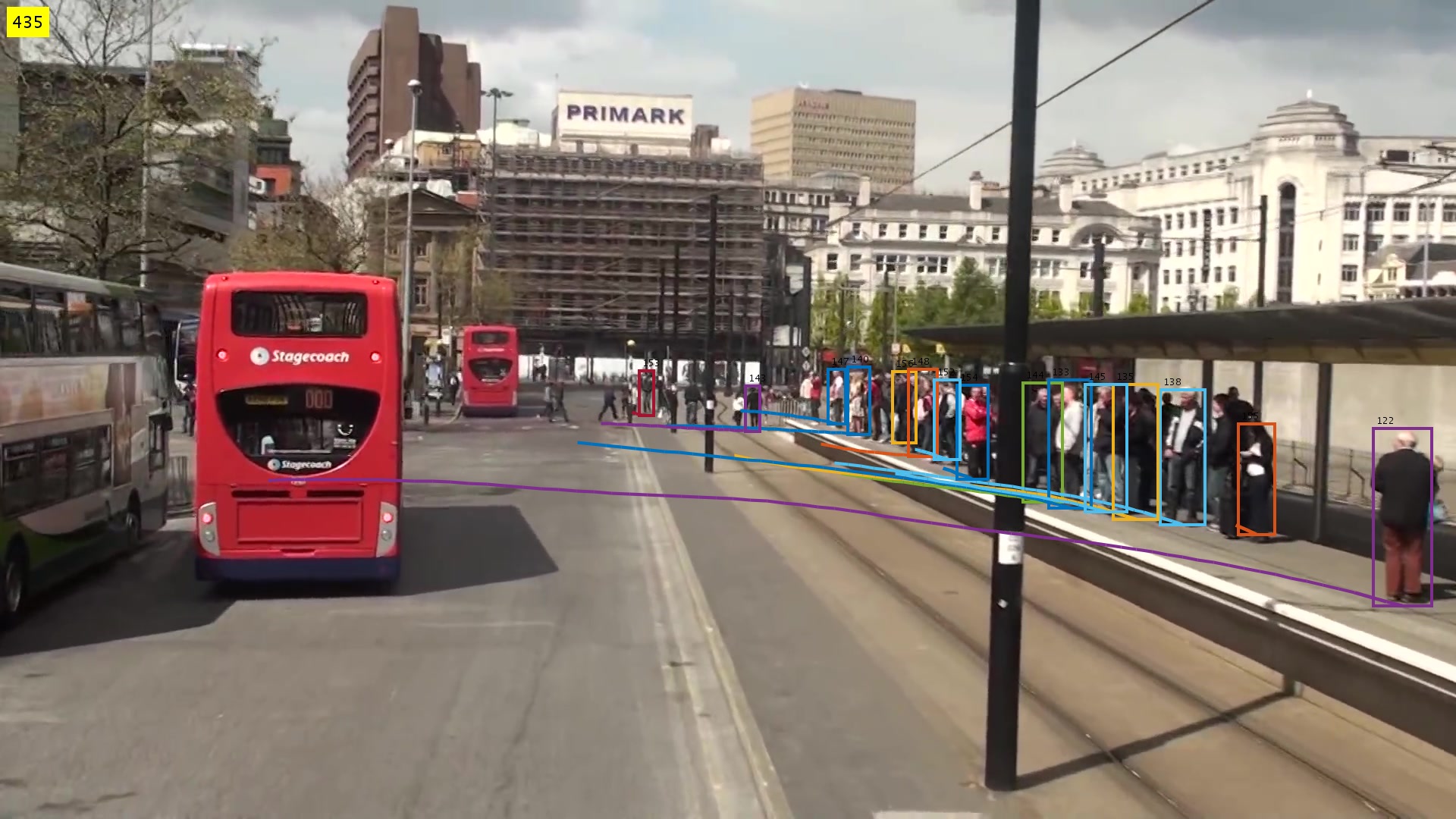}\\
		MOT16-06 & MOT16-03 & MOT16-12 & MOT16-14\\
		
	\end{tabular}
	\caption{ Visualization of selected sequences. The first row is from MOT15 test set, the second row is from MOT16 test set. Trajectories are fitted for better view.}
	\label{fig_results_vis}
\end{figure*}
\subsection{Ablation Study}

We justify the effectiveness of each building block in proposed method through ablation study as shown in Tab.~\ref{tab:ab}. IA stands for the proposed instance-aware tracker. MR is the dynamic model refreshing. IA$-$DV disables detection verification in IA tracker by setting $T_V \to \infty$, which removes the awareness between target models thus is equivalent with applying multiple independent SOT trackers for MOT. IA$-$TA disables target level awareness by replacing the fusion features with general deep features extracted from VGG-16 trained on ImageNet. Full method also includes the re-tracking and occlusion handle part.


Analysis is performed on Venice-2 sequence from training set of MOT15. Numerical results of all CLEAR MOT metrics are listed in Tab.~\ref{tab:ab}. Having demonstrated the importance of awareness between target models in Sec.~\ref{sec_tv}, totally disabling detection verification results in the greatest performance degradation. Model refreshment is also essential for improving performance and robust tracking. As shown by MOTA and MOTP, with model refreshment, not only tracking accuracy but also the bounding box precision improves a lot. In Full method, re-tracking and occlusion handle mechanism use simple linear interpolation to estimate the missing locations between the previous tracked target and current detection, which may introduce FP, but reduces more FN as shown in Tab.~\ref{tab:ab} thus still improves the overall performance.

\subsection{Results on Test Sequences}

We test our proposed approach on both MOT15 and MOT16 test sequences. In order to boost performance, we adopt several pre- and post- processing techniques, including excluding detections with extreme size according to scene prior and applying fitting to result trajectories in sequences where no rapid pedestrian scale changes. The performance is shown in Tab.~\ref{table:res15} and Tab.~\ref{table:res16}. We compared our method with the best peer-reviewed and published results on the benchmark, including JMC~\cite{keuper2016multi}, AP~\cite{long2017online}, LMP~\cite{tang2017multiple} and AMIR~\cite{sadeghian2017tracking}. 

The biggest challenge in MOT15 and MOT16 datasets is the enormous FN over FP (more than 10 times in MOT16) as shown in  Tab.~\ref{table:res15} and Tab.~\ref{table:res16}, which is partially introduced by FN in public detection. Benefited from the built-in SOT techniques, proposed method results in the least number of FN and the best MT/ML performance compared with all other online methods. As for overall performance, we established a new state-of-the-art among all online and offline methods in both MOT15 and MOT16 benchmark in terms of MOTA which is the most important metric for MOT. Visualization of selected sequences is shown in Fig.~\ref{fig_results_vis}. The complete metrics and visualization can be found on the benchmark website.\footnote{\url{https://motchallenge.net/results/2D_MOT_2015/} and \url{https://motchallenge.net/results/MOT16/} referred as `KCF'.}

\section{Conclusion}
\label{conclude}
In this paper we proposed using instance-aware with SOT technique to improve multiple object tracking (MOT). By built-in instance-awareness both in each target model and between all target models, our proposed approach can better predict the location of each target online, and meanwhile conserves the uniqueness of each tracking model to prevent the generation of duplicated and false positive trajectory. Tracking models in our approach are refreshed dynamically with a learned convolutional neural network to inhibit the noise of using inaccurate detections and to adapt appearance and scale variation of targets over time. Experiments on the MOT15 and MOT16 challenge datasets show the effectiveness of proposed approach in comparison with state-of-the-art. 

{\small
\bibliographystyle{ieee}

}

\end{document}